\documentclass[runningheads]{llncs}

\usepackage{eccv}

\usepackage{eccvabbrv}

\usepackage{graphicx}
\usepackage{booktabs}
\usepackage{colortbl}
\usepackage{multicol}
\usepackage{multirow}
\usepackage{amsmath}
\usepackage{tabularx}
\usepackage{algorithm}
\usepackage{algpseudocode}

\usepackage[accsupp]{axessibility}  

\usepackage{hyperref}

\usepackage{orcidlink}
\DeclareMathOperator*{\argmin}{arg\,min}
\DeclareMathOperator{\atantwo}{atan2}

\begin{document}

\title{StreamLTS: Query-based Temporal-Spatial LiDAR Fusion for Cooperative Object Detection} 

\titlerunning{StreamLTS}

\author{Yunshuang Yuan \orcidlink{0000-0001-5511-9082} \and
Monika Sester \orcidlink{0000-0002-6656-8809}}


\institute{Leibniz University Hannover, Germany\\
\email{\{firstname.name\}@ikg.uni-hannover.de}}

\maketitle

\begin{abstract}
  Cooperative perception via communication among intelligent traffic agents has great potential to improve the safety of autonomous driving. However, limited communication bandwidth, localization errors and asynchronized capturing time of sensor data, all introduce difficulties to the data fusion of different agents. To some extend, previous works have attempted to reduce the shared data size, mitigate the spatial feature misalignment caused by localization errors and communication delay. However, none of them have considered the asynchronized sensor ticking times, which can lead to dynamic object misplacement of more than one meter during data fusion. In this work, we propose Time-Aligned COoperative Object Detection (TA-COOD), for which we adapt widely used dataset OPV2V and DairV2X with considering asynchronous LiDAR sensor ticking times and build an efficient fully sparse framework with modeling the temporal information of individual objects with query-based techniques. The experiment results confirmed the superior efficiency of our fully sparse framework compared to the state-of-the-art dense models. More importantly, they show that the point-wise observation timestamps of the dynamic objects are crucial for accurate modeling the object temporal context and the predictability of their time-related locations. The official code is available at \url{https://github.com/YuanYunshuang/CoSense3D}.
  \keywords{Cooperative Perception \and Point Cloud \and Data Fusion}
\end{abstract}

\section{Introduction}
\label{sec:intro}

\begin{figure}[t]
  \begin{subfigure}{0.48\linewidth}
  \centering
    \includegraphics[width=0.8\textwidth]{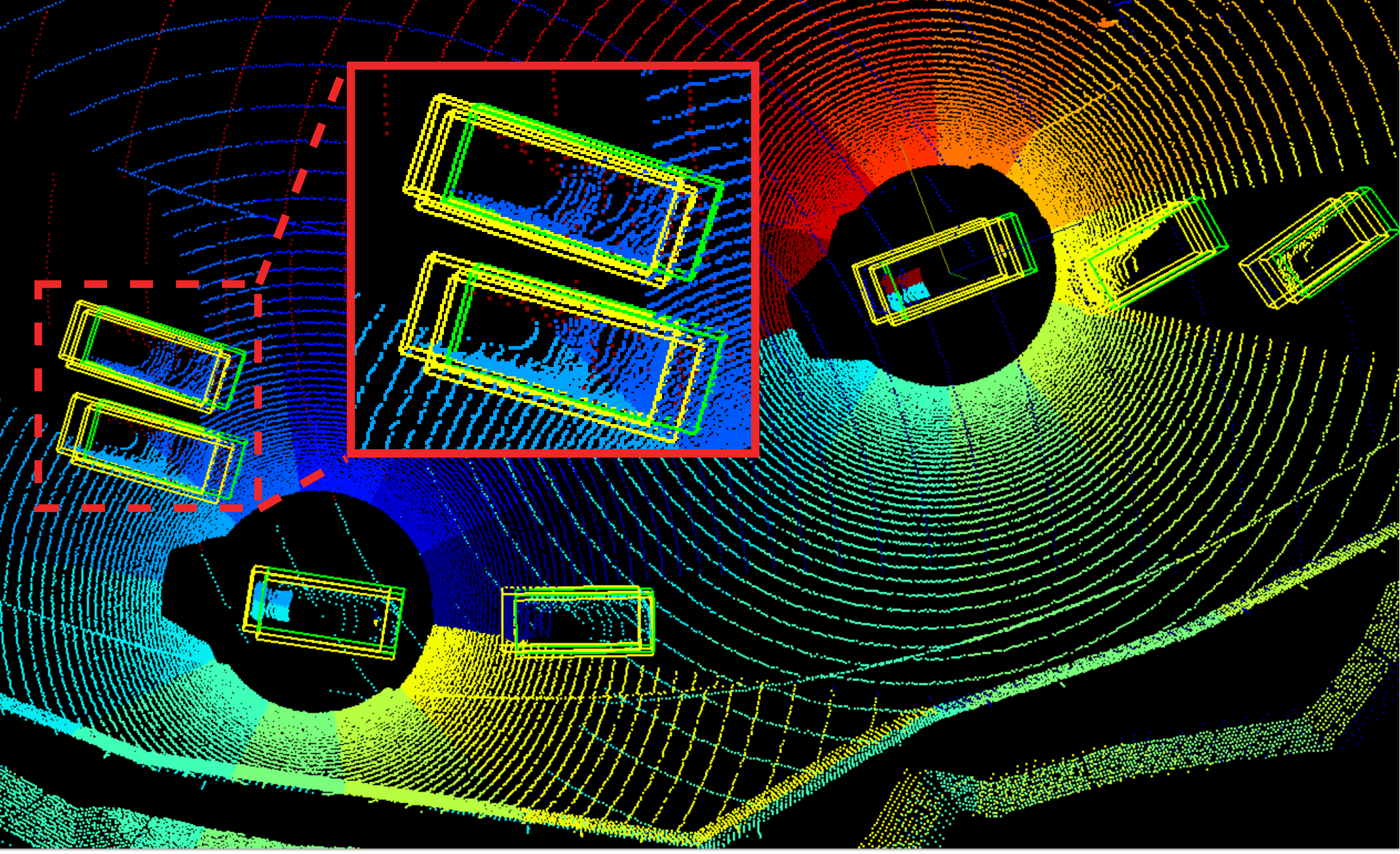} 
    \caption{OPV2Vt}
    \label{fig:opv2vt}
  \end{subfigure}
  \begin{subfigure}{0.48\linewidth}
  \centering
    \includegraphics[width=0.8\textwidth]{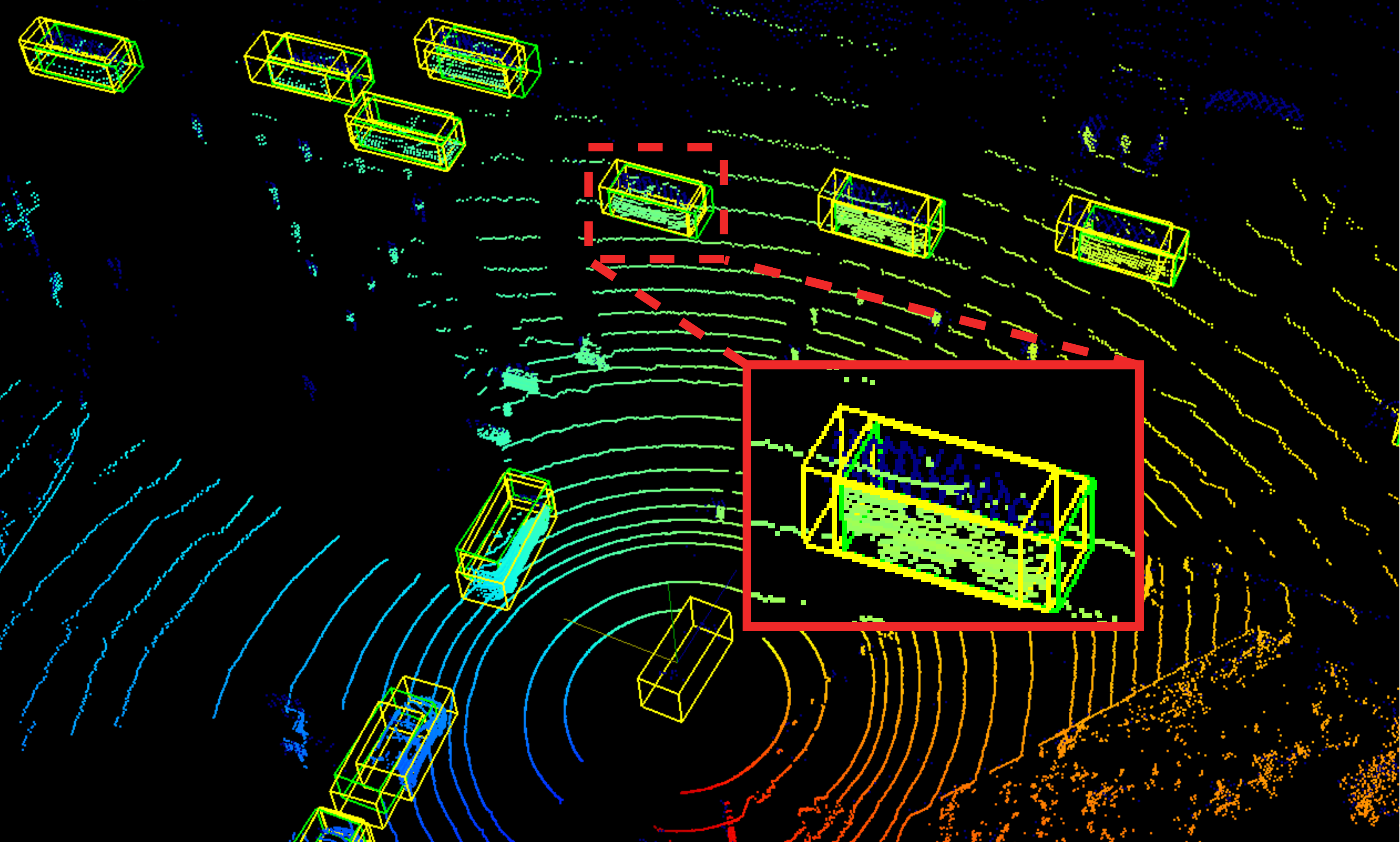} 
    \caption{DairV2Xt}
    \label{fig:dairv2x}
  \end{subfigure}
  \caption{TA-COOD: The scan points (blue to red scaled colors) and objects are observed at different timestamps, leading to unaligned object ground-truth (yellow boxes). The bounding boxes at the scan end of this frame are taken as the final ground-truth for TA-COOD (green boxes).}
  \label{fig:scenes}
\end{figure}

As the economy expands, the demand for mobility grows substantially, amplifying the complexity of transportation within constrained time and space. This surge poses significant challenges to developing effective traffic solutions without compromising safety and comfort. Autonomous Vehicles (AVs) emerge as one potential solution to improve road safety and efficiency. Perceiving the driving environment with the on-board sensors is crucial for autonomous driving. However, the perception performance is limited by the perceiving range of sensors and occlusions caused by ego-view sensor placement.

In recent years, cooperative perception with sharing information among Intelligent Agents (IAs), such as Connected Autonomous Vehicles (CAVs) and Connected Infrastructures (CIs), has proven to be beneficial for improving the perception accuracy\cite{Chen2019FcooperFB, Xu2022opv2v, fpvrcnn, Wang2020v2vnet, Yu2022dairv2x}. Different from the ego-perception system that only observes the driving environment from the ego-view, cooperative perception collect information from the sensors mounted on different IAs. These sensors typically have better spatial arrangement, can efficiently mitigate the occlusions and increase the perception distance of the ego vehicle. This ensures safer driving and more optimized route planning of AVs. However, sharing and fusing information from different IAs introduces several challenges, including the limited communication bandwidth and information misalignment introduced by localization errors and asynchronized capturing time of sensor data.

Previous works have sought to minimize bandwidth usage by compressing the learned deep Bird's Eye View (BEV) features~\cite{Wang2020v2vnet, Chen2019FcooperFB, Xu2022opv2v, TransIFF,fan2023quest} or selecting deep keypoint features~\cite{fpvrcnn}, rectify localization errors through learning-based techniques~\cite{Wang2020v2vnet, Xu2022v2xvit} or analytic algorithms~\cite{fpvrcnn, Yuan2022loccor}, and synchronize communication delays with the attention mechanism~\cite{Xu2023_v2vreal} or explicitly predicting the future features before data fusion~\cite{Yu2023v2xseq}. Nonetheless, none considered the asynchronized sensor ticking time. 

More specifically, previous benchmarks~\cite{Xu2022opv2v, Yu2022dairv2x, Xu2023_v2vreal, Yu2023v2xseq} assume that the sensors have synchronized ticking time in each aligned frame. In reality, sensors tick asynchronously, leading to inhomogeneous observation time of each object in the scenario. This effect can be amplified when the rolling shutter effect (\eg mechanic LiDARs) exists. At the sensor capturing frequency of 10Hz, these time differences ($0~\text{to}~100ms$), that are in the same magnitude as the communication delay, can lead to large spatial displacement at data fusion, especially when the CAVs have high speeds. Therefore, they should not be ignored. As the two example scenes shown in \cref{fig:scenes}, we leverage the timestamps of each point (illustrated with differently colored points) in the point cloud to achieve globally \textbf{T}ime-\textbf{A}ligned \textbf{CO}operative \textbf{O}bject \textbf{D}etection (\textbf{TA-COOD}) that predicts object bounding boxes (BBoxes) for the unified global timestamp (green boxes) in stead of detecting the timely asynchronous local bounding boxes (yellow boxes). 

To model more robust temporal features, we utilize the information of previous frames. This is very computational-demanding as the data from several IAs should be processed. Most previous works~\cite{Xu2022opv2v, Wang2020v2vnet, Yin2023v2vformer_plus, Xu2022v2xvit, He2023v2x_ahd} use dense convolutions or transformers to process BEV feature maps, that consume a lot of GPU memory and are hard to extend the model to sequential data on limited computational resources. In contrast, the potential of fully sparse network structures has rarely been researched. In this work, we built a fully sparse framework that can be trained on a single \textit{Nvidia GTX 3090} with sufficient batch size (\eg 4) and processing LiDAR streaming data of more than 4 frames and up to 7 CAVs.

We use fully sparse convolutions to encode the point cloud data and only select the keypoints in the Region of Interest (RoI) for further processing. 
Inspired by the query-based object detection~\cite{detr3d,Liu2022PETRv2AU,Wang2023streampetr}, we regard the RoI keypoints as queries of transformer input and follow StreamPETR~\cite{Wang2023streampetr} to model the temporal data fusion. 
To reduce the communication overhead, we fuse the temporal data before sharing to other IAs. More specifically, the selected local query points interact with both the historical query points in the previous frames and the keypoint features in the current frame to aggregate the temporal and spatial features for the ego-based time-aligned object detection. These aggregated query features are then shared and spatially fused by a sparse attention module for the TA-COOD task. Since our framework fuses both \textbf{T}emporal and \textbf{S}patial information from \textbf{Stream}ing \textbf{L}iDAR data of IAs, we call it \textbf{StreamLTS}.

To evaluate StreamLTS, we conduct experiments on the dataset DairV2X~\cite{Yu2022dairv2x} with adaptions to the TA-COOD problem based on the provided point-wise timestamps for each point cloud. Only with the cooperation between a vehicle and an infrastructure, this dataset has limited dynamics of sensors. Therefore, we simulate a new dataset with more CAVS and larger dynamics based on the scenario replay files of the dataset OPV2V~\cite{Xu2022opv2v}. Both datasets take ground-truth at global-aligned timestamp. Owning to the detailed temporal features, we notate the two datasets as OPV2Vt and DairV2Xt, respectively.

Our contributions are threefold. First, We proposed a TA-COOD that handles asynchronous data and aims to generate global aligned BBoxes for objects. To this end, we adapted the OPV2V and DairV2X datasets with considering the asynchronous ticking time of sensors to evaluate the models for TA-COOD.
Second, We developed a highly efficient and fully sparse framework \textbf{streamLTS} that fuses both temporal and spatial information from multiple IAs and generate the TA-COOD results.
Third, Our experimental results show that the sub-frame observation timestamps of the obtained data are crucial to improve the positioning accuracy of the detected objects at the given global aligned timestamp.

\section{Related Work}

\subsection{Cooperative Object Detection}
Object detection is crucial for autonomous driving to avoid collisions. By exchanging perceived objects, \cite{object_fusion} proved that sharing information among IAs significantly increases the detection accuracy compared to the ego-based perception. Instead, \cite{cooper, Marvasti2020CooperativeLO, Xu2022opv2v, Wang2020v2vnet} compared the performance of sharing and fusing data of different processing stages, including raw, semi- and fully-processed data fusion, all show that semi-processed data fusion has the potential to achieve the best performance with limited and controllable communication resource consumption. 

However, semi-processed data can be in any format.
For instance, Fcooper~\cite{Chen2019FcooperFB} experimented with both voxel feature and feature map fusion, found that voxel feature fusion has better performance as it keeps more details.  Instead, \cite{Xu2022opv2v} encodes the point clouds into point-pillars~\cite{pointpillar} with fully-connected layers and uses dense convolutions to generate feature maps of different IAs, and then fuse them with an attention module. These learnable encoding and fusion modules ensure that the most important information is selected and shared to achieve better performance.
To further increase the detection accuracy despite spatial misalignment of feature maps, SCOPE~\cite{scope} introduced the pyramid LSTM~\cite{Lei2022latency} to reason about temporal and spatial information from sequential point clouds. 

While these works all take feature maps as their fusion strategy, \cite{Where2comm} and  \cite{Yuan_gevbev2023} utilize the learned criterion, such as confidence and uncertainty, to select the most relevant information and reduce the data size for sharing. Similarly, \cite{fpvrcnn} adopt the sparsity of point clouds and learn the most important object keypoints for sharing. To further reduce the bandwidth requirement, \cite{TransIFF} selects the top ranking point features from the BEV feature map as object queries for further processing, sharing and fusing. They all achieve good performances with significantly less data sharing. 

Except \cite{Yuan_gevbev2023}, all above mentioned works encode point clouds into BEV feature maps and process them with dense convolutions or transformers which are very computationally demanding. The efficiency of the sparse operations are rarely explored. In this work, we design a fully sparse and training-efficient framework to model the spatio-temporal information from streaming LiDAR data.

\subsection{Query-based Temporal Data Fusion}
Based on transformer, DETR~\cite{detr} first introduced the query-based object detection for 2D images. Since it is anchor-free and needs no post-processing, it has been widely used and was extended to 3D object detection~\cite{detr3d, liu2022petr, Liu2022PETRv2AU}.   More importantly, the object queries provide great convenience to efficiently interact with features of any modality. For instance, by integrating the 3D geometry information into the image memory features, PETR~\cite{liu2022petr} built a simple yet effective framework to let 3D queries directly interact with the 2D features in image space and obtained superior 3D object detection performance. 
By propagating the learned object queries to the next frames and model the temporal context for the current frame, QueryProp~\cite{queryprop} improved the efficiency and accuracy for the video object detection. With only images as input,  StreamPETR~\cite{Wang2023streampetr} achieved on-par 3D object detection performance to the LiDAR-track benchmark on the NuScenes~\cite{nuscenes2019} dataset. Further more, the concept of object query propagation has been applied to object tracking tasks. MOTR~\cite{zeng2021motr} and TrackFormer~\cite{meinhardt2021trackformer} use object queries to achieve temporal modeling and associate the objects between frames in an end-to-end manner, which is simple yet effective on tracking.
QUEST~\cite{fan2023quest} extended PETR to cooperative object detection without considering the temporal information.
In this work, we model the temporal context of observations from different IAs with the selected potential object queries, and share the learned queries for TA-COOD to save communication bandwidth.

\subsection{Datasets for Collective Perception}
It is highly expensive and complicated to obtain training data for cooperative object detection. It requires several vehicles and sensors observing the same scene. This scale leads to highly complex calibration and post-processing to generate accurate ground-truth meta information (\eg sensor poses) and annotations. Therefore, the initial attempts for collective perception either use simplified driving scenarios~\cite{cooper, Chen2019FcooperFB} or synthetic data~\cite{Xu2022opv2v, Yuan2021comap, Li2022v2xsim}. For instance, \cite{Chen2019FcooperFB} conducted their cooperative perception experiments on a dataset captured from a static parking lot with static sensors. This is the simplest set up without considering any dynamics of sensors or objects in the scenario. Via simulation, \cite{Xu2022opv2v} generated a large-scale dataset for collective perception and built a cooperative object detection benchmark, however, without considering sensor asynchrony. 

Recently, the real datasets DairV2X~\cite{Yu2022dairv2x} and V2V4Real~\cite{Xu2023_v2vreal} were made accessible. Both are only configured with two agents, DairV2X with one CAV and one connected infrastructure, V2V4Real with two CAVs. Because of the dynamics and asynchrony of sensors, there might be spatial misalignment between dynamic objects observed by the two agents. To generate unified ground-truth BBoxes for cooperative object perception, both DairV2X and V2V4Real take ego-vehicle's annotation as the ground-truth if the annotations from two agents spatially overlap. However, at the blind spot of the ego-vehicle, the annotations of the cooperative agent are taken. In this way, the final generated bounding box might contain errors and does not reflect the true location of the encapsulated object at the given timestamp. This might also negatively influence the temporal predictability of sequential models. To better preserve the detailed temporal information and its relation to the accurate location of the objects in the scenario, we proposed to generate global time-aligned ground-truth BBoxes for the TA-COOD task.

\section{Methodology}
\subsection{Definition of Time-Aligned Cooperative Object Detection}

\begin{figure}[t]
  \centering
    \includegraphics[width=0.5\textwidth]{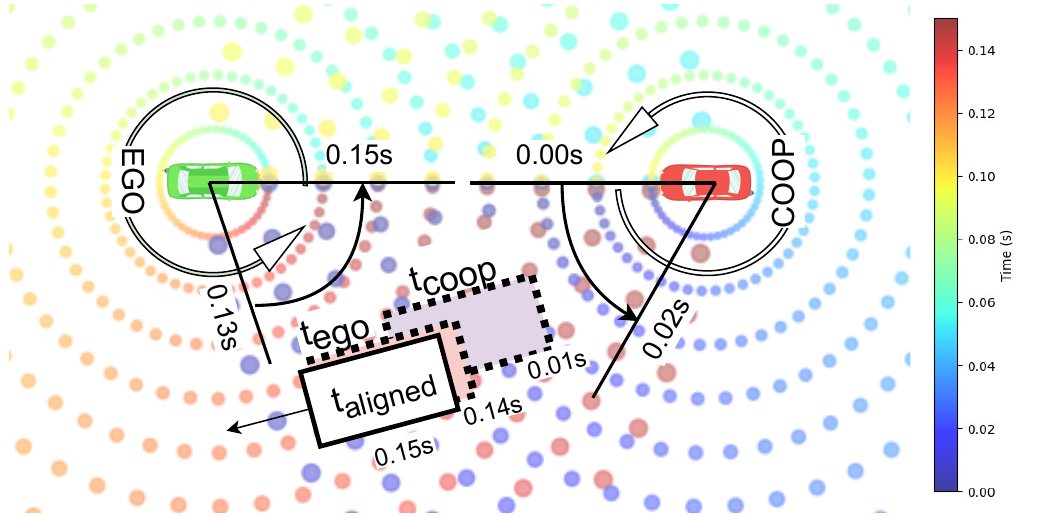} 
  \caption{TA-COOD with two CAVs. Dashed boxes: ground-truth BBox of ego- and coop-CAV observations. Solid box: time-aligned ground-truth BBox.}
  \label{fig:temp_det}
\end{figure}

In a cooperative perception scene, assume that a set of IAs $\textbf{A}=\{A_i |0 \leq i \leq N\}$ are inter-connected, where $A_0$ is the ego IA. Each IA has its own sensor ticking time $t_i$. All sensors are scanning at the frequency $f$. An example with two CAVs is shown in \cref{fig:temp_det}. The LiDAR of the ego IA $A_0$ (green) and the cooperative IA $A_1$ (red) are scanning counterclockwise. They have the ticking frequency of $f=10Hz$ and the ticking time offset of $t_0 - t_1 = 0.05s$. The scanning time of each observation point is shown with gradient colors. As shown in the middle-bottom area of \cref{fig:temp_det}, the observation by the ego- and the coop-vehicle in this area have a time difference of $0.11s \leq t_{ego} - t_{coop}\leq 0.15s$. At the speed of $60km/h$, this time offset can lead to the target object shift of more than $1.8m$, introducing difficulty into the data fusion of the IAs. Instead of taking the annotation from one IA (\eg the red BBox observed by the ego IA at $t_{ego}$), we generate the ground-truth BBox at the globally aligned time $t_{aligned}$ by interpolating between the annotations of the individual CAVs. To accurately model the movement of the detected object, we define that the 3D TA-COOD aims to take point-wise timestamped point clouds as input and predict the bounding box at globally aligned time $t_{aligned}$ which is the scan end of the reference vehicle in each frame. Since we evaluate the detection result in the perspective of ego vehicle, therefore we take ego vehicle as reference. For the example in \cref{fig:temp_det}, the global time is aligned to $t_{aligned}=0.15s$, the scan end of the ego-vehicle.

\subsection{Overall Framework}
\begin{figure}[t]
  \centering
    \includegraphics[width=0.95\textwidth]{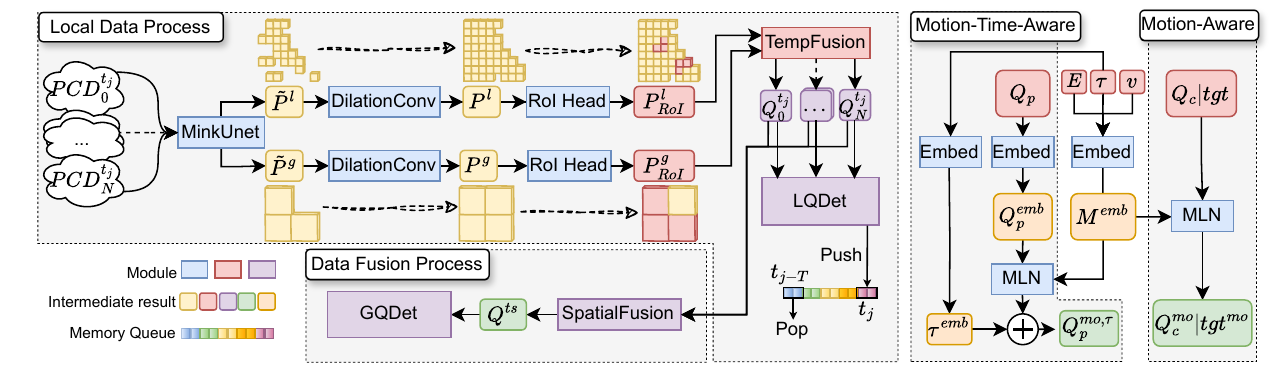} 
  \caption{Left: Overall framework for TA-COOD. \textit{Local Data Process} encodes local point cloud features and fuse the temporal context into the object queries; Weights are shared across agents. \textit{Data Fusion Process} fuses the spatial object query features from the $N$ IAs. Right: MTA alignment. \textit{Motion-Time-Aware} embeds the query positions ($Q_p$) and \textit{Motion-Aware} embeds the query context $Q_c$ and initial queries ($tgt$).}
  \label{fig:streamLTS}
\end{figure}

The overall framework of processing one frame data in a sequence is shown in left part of \cref{fig:streamLTS}. 
Given the point clouds $PCD_0^{t_j},\dots, PCD_N^{t_j}$ from $N$ IAs at time $t_j$, a shared \textit{MinkUnet}~\cite{choy20194d, Yuan_gevbev2023} encodes them into multi-resolution sparse features, including the local feature at higher resolution $\tilde{P}^l$ and the global feature at lower resolution $\tilde{P}^g$. They are then dilated by \textit{DilationConv} module, separately, to extend their coordinate coverage. The dilated features $P^l$ and $P^g$ are then fed to the corresponding \textit{RoI Head} to generate importance scores which are used for selecting the top-$K_{roi}$ RoI points $P^l_{RoI}$ and $P^g_{RoI}$. The following shared \textit{TempFusion} module is composed with a Transformer decoder which takes $P^l_{RoI}$ as new object queries to interact with the stored historical queries from $T$ previous frames to obtain temporal context and with the global feature $P^g_{RoI}$ to obtain spatial context at the current frame. On the one hand, the outputs $Q_0^{t_j}, \dots, Q_N^{t_j}$ of \textit{TempFusion} are fed to a local query-based detection head \textit{LQDet} to obtain the detection result at each IA. Based on the detection scores, only the top-$K_q$ object queries are pushed into the memory queue while the queries at the oldest timestamp $t_{j-T}$ are discarded. On the other hand, these outputs are shared to the ego IAs for \textit{Spatial Fusion} to obtain the temporal- and spatial-aligned features $Q^{ts}$. Based on these features, the global detection head \textit{GQDet} generates the TA-COOD result after data fusion.

\subsection{Module Configuration}
\textbf{Backbone: MinkUnet and DilationConv} 

\textit{MinkUnet} is composed by fully sparse convolution layers which is highly efficient to encode sequential point cloud data. The point clouds are voxelized with the size $0.4m$ and embedded to voxel features with a MLP layer. Then the 3D sparse convolution layers are stacked in the Unet~\cite{unet} structure to encode the voxel features. The encoded features are then compressed along the height dimension by 3D sparse convolution layers into sparse 2D BEV features. We take the BEV feature at down-sample level 2 ($0.8m$ per pixel) as $\tilde{P}^l$ and level 8 ($3.2m$ per pixel) as $\tilde{P}^g$. \textit{DilationConv} has three sparse convolution layers with coordinates expansion and kernel size 3. At each layer, the coordinates of sparse tensors are dilated for one pixel. 

\noindent\textbf{Heads: RoI and QDet} 

\textit{RoI Head} aims to select the most interesting features for further processing. It uses linear layers to generate importance scores. For $P^g$, we expect that the importance of input features should be learned automatically by \textit{TempFusion}. Therefore, no loss for this head. For $P^l$, we use a center-based~\cite{yin2021center} detection head which uses the Focal loss~\cite{focal_loss} of classification to learn importance scores and the Smooth L1 loss of regression to help the backbone network to learn the geometric features of the objects.
The \textit{Query-guided Detection head (QDet)} heads, including the Local QDet (LQDet) in \textit{TempFusion} and the Global QDet (GQDet) in \textit{SpatialFusion}, have the same structure as the \textit{RoI Head} for $P^l$. They generate TA-COOD results with respect to the query positions and features. 

\noindent\textbf{TempFusion} 

\begin{figure}[t]
  \centering
    \includegraphics[width=0.98\textwidth]{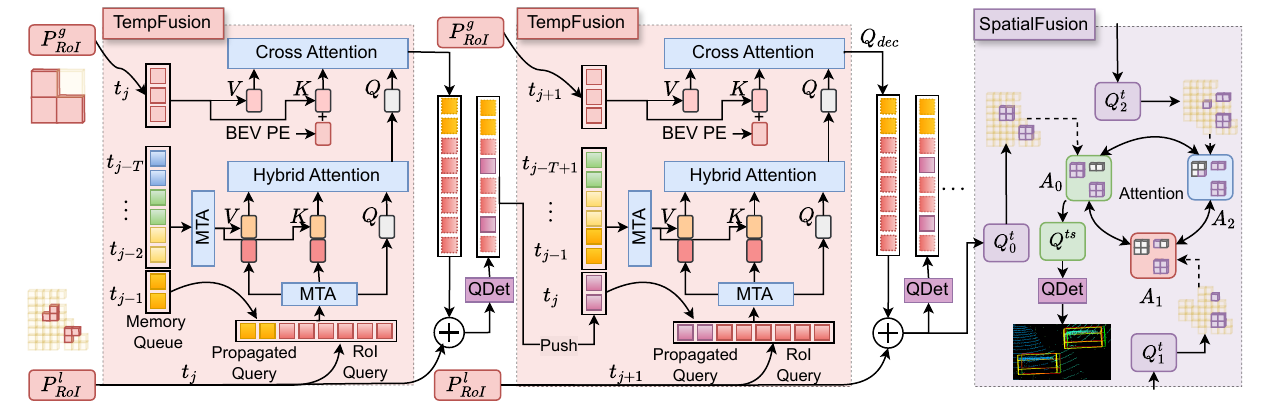} 
  \caption{The details of the \textit{TempFusion} and \textit{SpatialFusion} module}
  \label{fig:fusion}
\end{figure}

\textit{TempFusion} is built based on the StreamPETR~\cite{Wang2023streampetr}. The detailed workflow of two consecutive frames is shown in \cref{fig:fusion}.
It contains four components: \textit{Memory Queue}, \textit{Motion-Time-Aware (MTA)}, \textit{Hybrid Attention} and \textit{Cross Attention}. 
The \textit{Memory Queue} stores the top-$K_q$ interesting object queries from each of the $T$ historical frames. 
At the aligned time $t_j$, the input feature $P^l_{RoI}$ is extended by the propagated top-$K_q$ memory object queries at time $t_{j-1}$. The \textit{Hybrid Attention} uses these queries to interact with all queries from the current frame $t_j$ and the memory queue of frame $t_{j-1} \dots t_{j-T}$ to learn temporal context of each object. However, before this attention, all query, key and value features are spatially and temporally aligned to the current pose and time using \textit{MTA}. 
The output of the \textit{Hybrid Attention} is then used as queries of the \textit{Cross Attention} to interact with the global feature $P^g_{RoI}$ to obtain spatial context in the current frame. The output queries $Q_{dec}$ are added with the initial RoI features $P^l_{RoI}$ to enhance the local context in the current frame. The summation $Q_i^{t_j}$ are then fed to the \textit{LQDet} head to generate the time-aligned objects locally at each IA. Based on the detection scores, the top-$K_q$ object queries are then pushed into the memory queue for the next iteration of temporal fusion at time $t_{j+1}$ as shown in the middle of  \cref{fig:fusion}.

\textbf{Memory queue}: stores $K_q$ object queries for each of the $T$ previous frames. For each selected object query, the predicted object center $Q_p$, the context embedding $Q_c$, the velocity $v$, the pose matrix $E$ and the query timestamp $\tau$ of the corresponding IA are stored. The timestamp $\tau$ is retrieved following \cref{eq:tau}, where $Q_x$ and $Q_y$ are the $x$ and $y$ coordinate of the query point, $P_x^i$, $P_y^i$ and $P_t^i$ are respectively the $x, y$ coordinates and the timestamp of the $i$th point $P^i$ in the point cloud.
\begin{equation}\label{eq:tau}
    \tau = P_t^i, \quad i=\argmin |\atantwo (Q_y, Q_x) - \atantwo (P_y^i, P_x^i) |
\end{equation}

\textbf{MTA}: is shown in the right of \cref{fig:streamLTS}. The timestamp $\tau$ of each query, the pose $E$ and the velocity $v$ of the corresponding ego-vehicle, are aligned as a vector and embedded to the feature $M^{emb}$. With the Motion-aware Layer Normalization (MLN~\cite{Wang2023streampetr}), this feature is then injected into the embedded query position $Q_p^{emb}$, the query context $Q_c^{emb}$ of previous frames, and the initial query target $tgt$ at the current frame to obtain the motion-aware embeddings. Additionally, the embedded time is added to the motion-aware query position embedding ($Q_p^{mo}$) to make it also time-aware ($Q_p^{mo,\tau}$).

\textbf{Difference to StreamPETR}: we made several changes to match the new data modality. Instead of frame-wise timestamp used in StreamPETR, we leverage the observation timestamp of each individual object query to capture the accurate temporal context. Due slow convergence of the StreamPETR on our datasets, we made three changes to accelerate training process. First, we take the top-$K_{RoI}$ weighted keypoints $P^l_{RoI}$ as the initial object queries instead of the random reference points. Second, we additionally add the initial RoI features $P^l_{RoI}$ to the decoded queries $Q_{dec}$ to enhance the local context so that the detection head can converge faster. Third, the DETR\cite{detr} detection head of StreamPETR is replaced by a center-based~\cite{yin2021center} detection head since the initial RoI queries are already good proposal center points for the BBox regression task. Only a refinement is required based on these center points. Additionally, we only use one decode layer in the transformer due to limited training resource.

\noindent\textbf{Spatial Fusion}

After each iteration of \textit{TempFusion}, the \textit{SpatialFusion} module fuses the local object queries $Q_i^t$ from different IAs . As the example of three IAs ($A_0, A_1, A_2$) shown in the purple block of \cref{fig:fusion}, the input query points $Q_i^t$ are firstly spatially aligned by taking the union of all query locations $P_i^t$ to obtain the same number of queries $\breve{P}_i^t$ at each IA. The query features at each padded location (gray cubes in \cref{fig:fusion}) are initialized with zeros (\cref{eq:q_pos_union}). These queries are then used for the global TA-COOD.
\begin{align}
    & \breve{P}_i^t = \bigcup_{i=0}^{N} P_i^t, \quad \breve{Q}_i^t = Padzero(Q_i^t, \breve{P}_i^t - {P}_i^t)
    \label{eq:q_pos_union}\\
    & Q = \breve{Q}_0^t, \quad K = V = Cat(\breve{Q}_0^t, ..., \breve{Q}_N^t) 
    \label{eq:qkv}\\
    & Q^{ts} = softmax(\frac{(Q \cdot K^T)}{\sqrt{d}}) \cdot V
    \label{eq:attn}
\end{align}

\section{Experiment}

\subsection{Datasets}
To evaluate the our proposed framework on the TA-COOD task, we generate two new datasets OPV2Vt and DairV2Xt based on the widely-used dataset OPV2V~\cite{Xu2022opv2v} and DairV2X~\cite{Yu2022dairv2x}, which are curated specifically for cooperative object detection that does not consider the sub-frame time asychrony of sensors. 

\textbf{OPV2Vt} (\cref{fig:opv2vt}): The orignal OPV2V dataset is simulated by CARLA~\cite{carla}. It contains about 12K frames of data in 73 scenes. We interpolate the meta data of OPV2V to generate ten intermediate locations for each object and sensor inside one frame, which leads to ten sub-frames (0.01s) in each scan frame (0.1s). We then replay all sub-frames to generate data with more fine-grained dynamics. Since these sub-frames are synchronized, we randomly drop the data of several beginning sub-frames for each scan sequence and joint every consecutive 10 sub-frames as a full LiDAR scan to simulate the ticking time asynchrony. For each jointed frame data, we use the BBoxes of the last sub-frame as ground-truth. More details can be found in Appendix A. One frame of data is shown in \cref{fig:opv2vt}. The evaluation detection range is set to $[-140.8, 140.8]m$, $[-38.4, 38.4]m$, and $[-3.0, 1.0]m$ for x-, y- and z-axis, respectively.

\textbf{DairV2Xt} (\cref{fig:dairv2x}): The original DairV2X is captured with one CAV and one CI at intersections. To generate the globally time-aligned BBoxes, we first refine the point cloud alignment between the CAV and the CI to decouple the BBox misalignment caused by localization errors and asynchronized capturing times. In each sequence, we select the point cloud of the CAV in the sequence where the location of the CAV is closest to CI as the global reference $O_{ref}$, and then register sequentially the point clouds of both CAV and CI in other frames to $O_{ref}$.  With the corrected sensor poses, the annotated objects are transformed from the local sensor coordinate to the globally aligned coordinate $O_{ref}$ and frame-wisely aligned to the object trajectories. By interpolation, we generate the ground-truth BBoxes that are aligned to the scan end $t_{aligned}$ in each frame. More details can be found in Appendix B. The detection range at y- and z-axis is the same as OPV2Vt. The range for x-axis is set to $[-102.4, 102.4]m$.

\subsection{Experiment Setting}
\textbf{Comparative experiment}: We select three state-of-the-art (SOTA) cooperative object detection frameworks, Fcooper~\cite{Chen2019FcooperFB}, AttnFusion~\cite{Xu2022opv2v} and FPVRCNN~\cite{fpvrcnn}, to conduct comparative experiments with our StreamLTS. 
To make a fair comparison, we attach our \textit{TempFusion} module to each framework to model the temporal interaction. For Fcooper and AttnFusion, we use the intermediate BEV features at the corresponding resolution as $P^l$ and $P^g$. FPVRCNN uses smaller input voxel size, leading to intermediate BEV features of higher resolutions. Therefore, we use two additional convolution layers to down-sample the BEV features to obtain the global features $P^g$. For the local features, we directly use keypoints generated by the Voxel-Set-Abstraction~\cite{Shi2019PVRCNNPF}. At \textit{SpatialFusion} stage, we use \textit{Maxout}~\cite{Chen2019FcooperFB} fusion for Fcooper and \textit{Attention} for the other two frameworks. We replace the dense convolutions of AttnFusion~\cite{Xu2022opv2v} with sparse convolutions to reduce memory usage. 

\textbf{Training}: We use the sliding window strategy to load the sequential data with 4 frames. For each training batch, we only calculate the loss for the newest frame to reduce the training time and GPU memory consumption. From each frame, we select $1024$ top-$k$ RoI keypoints from $P^l$ and $512$ top-$k$ global features from $P^g$ for \textit{TempFusion}, in which $K=256$ of the fused object queries are propagated to the next frame. To improve efficiency, we only calculate gradients for the ego-agent in the \textit{Local Data Process} (\cref{fig:streamLTS}) of OPV2Vt dataset. Gradients are calculated for both CAV and CI of DairV2Xt as they have different view angles, hence different spatial features. We augment both datasets by transforming the point clouds with a rotation angle in the range of $\pm \alpha ^{\circ}$, flipping along x- and y-axis and scaling  with the ratio in range $[0.95, 1.05]$. $\alpha$ is set to $90^{\circ}$.
Additionally, we follow \cite{Yuan_gevbev2023} to augment the input point cloud with free space points for StreamLTS to enhance the spatial connectivity of sparse tensors. To make the model also able to handle the large frame-wise time latency, we load cooperative data from a random previous frame ($t\in\{t_j, t_{j-1}, t_{j-2}\}$) instead of the aligned frame $t_j$. In general, we train all models for $50$ epochs with the Adam optimizer, batch size four for OPV2Vt and two for DairV2Xt. The learning rate is set to $2e-4$ with a warmup stage of $4000$ iterations.

\section{Experimental Results and Evaluation}

\subsection{Qualitative Results}
\begin{figure}[t]
  \centering
  \begin{subfigure}{0.48\linewidth}
    \includegraphics[width=\textwidth]{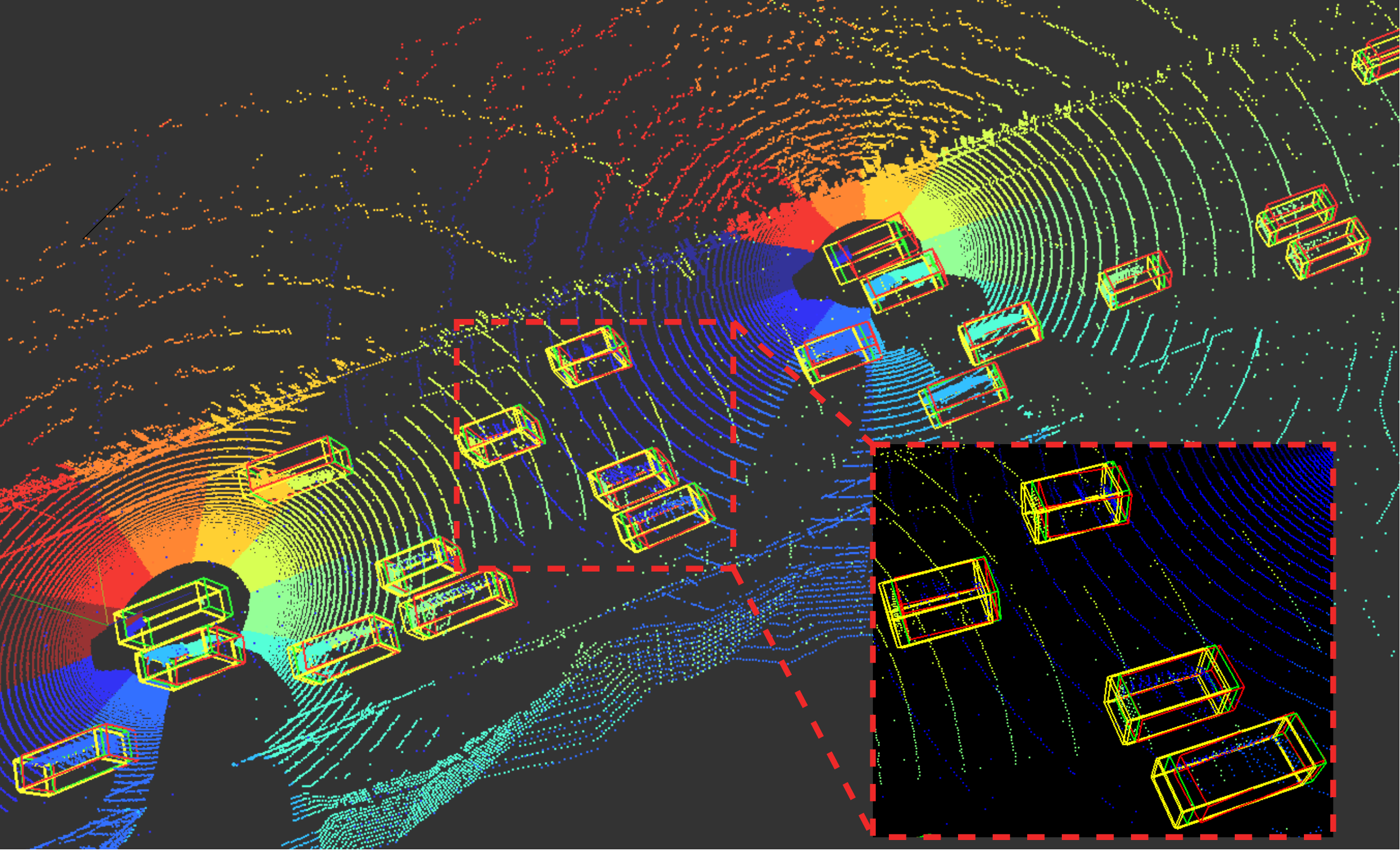} 
    \caption{OPV2Vt}
    \label{fig:res_opv2vt}
  \end{subfigure}
  \hfill
  \begin{subfigure}{0.48\linewidth}
    \includegraphics[width=\textwidth]{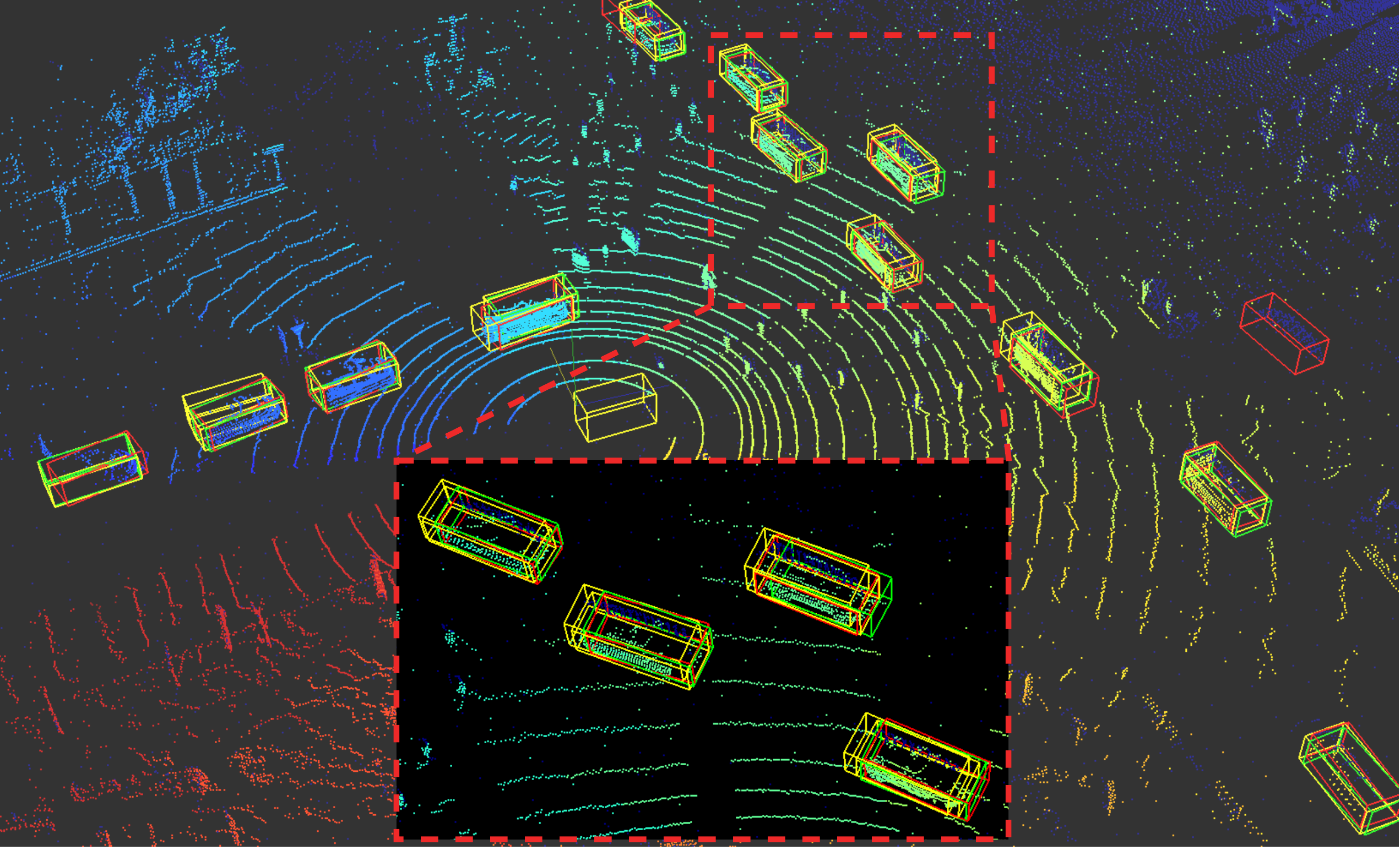} 
    \caption{DairV2Xt}
    \label{fig:res_dairv2x}
  \end{subfigure}
  \caption{TA-COOD result of StreamLTS. Yellow BBoxes: ego-view annotation. Green BBoxes: time-aligned annotation. Red BBoxes: time-aligned detection. }
  \label{fig:res_vis}
\end{figure}

TA-COOD result of StreamLTS is shown in \cref{fig:res_vis}. The yellow BBoxes are the ground-truth BBoxes at different observation times, the green BBoxes are the globally time-aligned ground-truth. As shown in the middle dashed red box of \cref{fig:res_opv2vt}, the objects in this area are observed by both CAVs at different times. The top right CAV scans this area (blue points) about $0.05s$ earlier than the bottom left CAV (yellow points). The global BBoxes are aligned to the scan end in this frame (same to the timestamps of the red points), hence moved  a little bit forward in comparison to the yellow BBoxes. This can also be observed in \cref{fig:res_dairv2x}. The red BBoxes are the predictions generated by StreamLTS. The result on the OPV2Vt and DairV2Xt datasets show that the predictions have better matches to the globally time-aligned ground-truth BBoxes (green) than any locally annotated BBoxes (yellow). This reveals that our framework has successfully captured the time relationships between the object observations from different IAs and correctly predicted the BBoxes at the aligned future timestamp.

\subsection{Comparison to SOTA Models}

\begin{table}[tb]
\caption{Average Precision (AP) of Time-aligned cooperative object detection.}
\centering
\begin{tabular}{l|l|l|cc|cc}
\toprule
\multirow{2}{*}{Framework} & \multirow{2}{*}{Backbone} & \multirow{2}{*}{Fusion} & \multicolumn{2}{c|}{OPV2Vt} & \multicolumn{2}{c}{DairV2Xt}\\ 
 &   & & \multicolumn{1}{c}{AP@0.5}    &  AP@0.7   & \multicolumn{1}{c}{AP@0.5} &  AP@0.7\\ \hline
Fcooper & PointPillar & Maxout  & \multicolumn{1}{c}{54.4}    &  17.0   & \multicolumn{1}{c}{41.8} &  17.7\\ 
FPVRCNN & Spconv & Attn & \multicolumn{1}{c}{70.8}    &  41.2   & \multicolumn{1}{c}{51.8} &  23.9\\ 
AttnFusion & VoxelNet & Attn & \multicolumn{1}{c}{\underline{78.7}}    & \underline{41.4} & \multicolumn{1}{c}{\underline{62.1}} &  \underline{34.0}\\ 
StreamLTS & MinkUnet & Attn & \multicolumn{1}{c}{\textbf{85.3}}    &  \textbf{72.1}  & \multicolumn{1}{c}{\textbf{64.0}} &  \textbf{40.4} \\ \bottomrule
\end{tabular}
\label{tab:compare_result}
\end{table}

\begin{figure}[t]
  \centering
    \includegraphics[width=0.85\textwidth, trim=2cm 0cm 2cm 0cm]{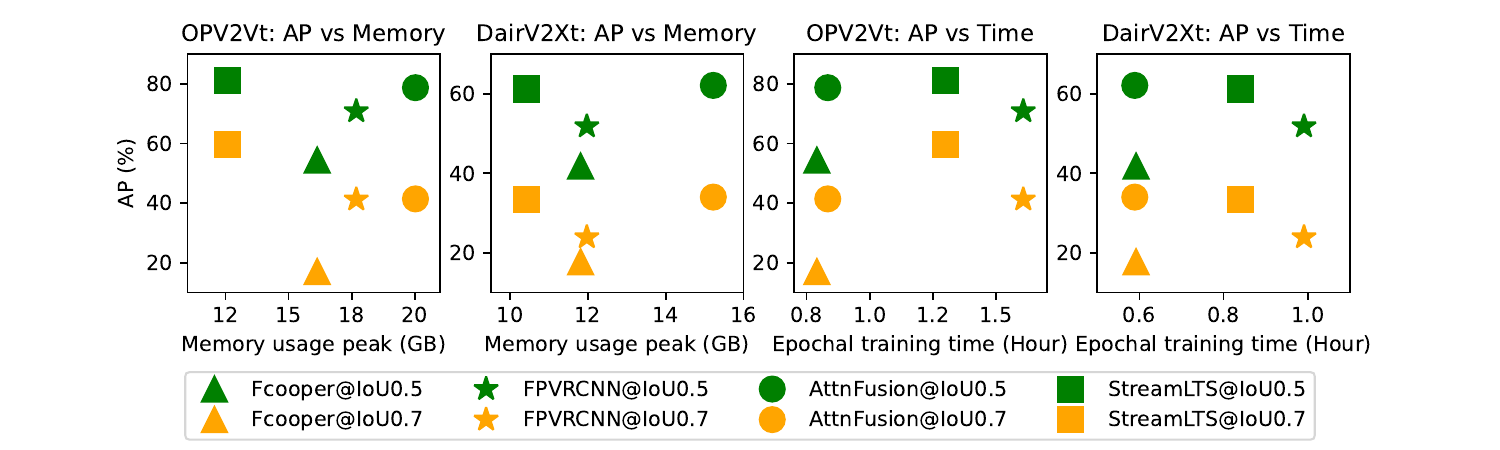} 
  \caption{Comparison of memory usage peak and epochal training time.}
  \label{fig:efficiency}
\end{figure}

\textbf{Average Precision (AP):}
As shown in \cref{tab:compare_result}, our framework StreamLTS achieves the highest AP on both datasets. Especially at the IoU threshold of $0.7$, the AP of our framework has increased $30.7\%$ on OPV2Vt and $6.4\%$ on DairV2Xt compared to the second-best framework AttnFusion.
Among all comparative frameworks, Fcooper has the worst performance as it uses a simpler backbone network that only consists of a PointPillar encoder and several 2D convolution layers, to encode the point cloud features. More importantly, its \textit{Maxout} operation for fusing the features from different IAs is not learnable, hence not able to handle the spatial misalignment of the object queries caused by different observation times. In comparison, all other three frameworks with enhanced backbone encoding and the learnable attention mechanism for multi-IA feature fusion have much better performance. Because of the high reliability on the accuracy of the first stage detection, FPVRCNN might ignore some potential object points for the next processing steps, hence achieves worse performance than AttnFusion and StreamLTS.

\textbf{Training efficiency:}
Less memory consumption and training time is crucial for faster model development on limited computing resources.
To investigate the training efficiency of StreamLTS, we compare it to other models with the peak memory usage because it is crucial for determining the devices the model can be trained on and the training batch size on limited GPU resource. In addition, we also tested the epochal training time of all comparative models on the same device. 
The results of AP against these two measures are demonstrated in \cref{fig:efficiency}. As shown in the left two sub-plots, our proposed StreamLTS (squares) requires the least training memory resource and has the best overall TA-COOD performance. AttnFusion has a comparable detection performance, however, its GPU memory demand is more than $50\%$ higher than StreamLTS. Regarding the epochal training time, the sparse models StreamLTS and FPVRCNN seem less efficient than the other two dense models. This is because the sparse operations are less parallelized for computation. However, thanks to the low memory usage, StreamLTS can reduce the training time by increasing the batch size.

\begin{figure}[t]
    \centering
    \includegraphics[width=0.92\textwidth]{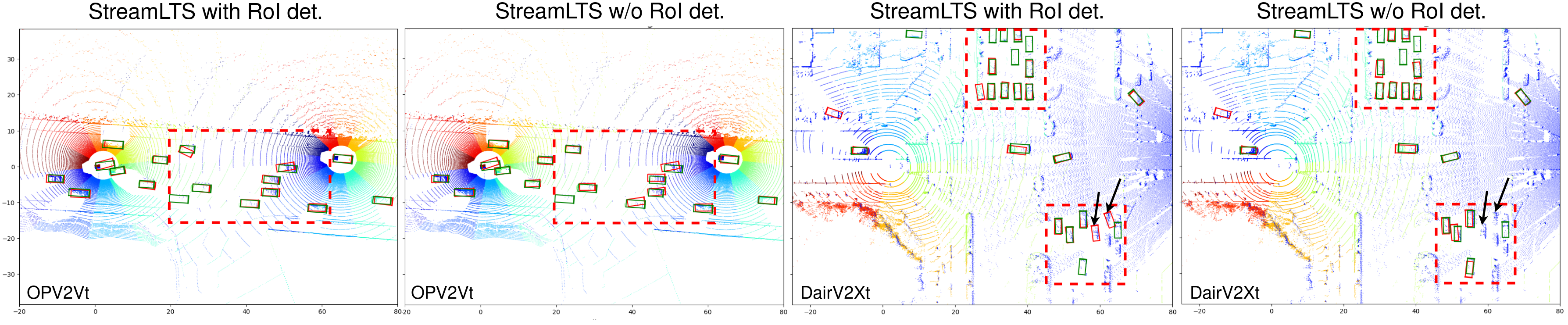}
    \caption{TA-COOD result of StreamLTS with and without RoI regression}
    \label{fig:ablation}
\end{figure}

\begin{table}[tb]
\caption{Ablation study on individual modules of streamLTS.}
\centering
\begin{tabular}{c|c|c|c|c|c|cc|cc}
\toprule
\multirow{2}{*}{Exp.} & Pt.& Temp.  & Dil. & RoI & Global & \multicolumn{2}{c|}{OPV2Vt} & \multicolumn{2}{c}{DairV2Xt}\\ 
 &Time & Fusion &  Conv & reg.  & attn &\multicolumn{1}{c}{ AP@0.5}    &  AP@0.7   & \multicolumn{1}{c}{AP@0.5} &  AP@0.7\\ \hline 
1 & &  & \checkmark & \checkmark& \checkmark&  77.4  &  53.3 & 55.4 & 34.0\\ 
2 & & \checkmark & \checkmark    &\checkmark & \checkmark& 73.4 & 38.7 & 56.4 & 36.7\\ 
3 & \checkmark &  \checkmark  &  & \checkmark  &\checkmark & 79.3 & 59.6 & {54.9} & {29.9} \\
4 &\checkmark & \checkmark & \checkmark    & & \checkmark & {83.6}   &  {69.7} & {60.2} & {34.4} \\
5 & \checkmark & \checkmark & \checkmark    & \checkmark & &  \underline{85.1}   &  \underline{71.0}& \underline{63.8} & \textbf{41.0}\\
6 & \checkmark & \checkmark & \checkmark    & \checkmark & \checkmark&  \textbf{85.3}   &  \textbf{72.1}& \textbf{64.0} & \underline{40.4}\\
\bottomrule
\end{tabular}
\label{tab:ablation}
\end{table}
\subsection{Ablation Study}
To validate the effect of individual modules on the performance, we conduct ablation study over the main modules of streamLTS. The results are shown in \cref{tab:ablation}. 
Experiment 1 and 2 validate the effectiveness of time-related modules on capturing the temporal context to predict more accurate BBoxes. However, using \textit{TempFusion} by replacing the point-wise time embedding of object queries with the unified frame-wise timestamps (Exp. 2) performs even worse than the model without this module (Exp. 1) on OPV2Vt. This reveals that the temporal modeling with inaccurate frame timestamps might be misleading. With the accurate point-wise timestamp (Exp. 6), the AP is nearly doubled. This confirms the effectiveness and importance of utilizing the accurate point-wise timestamps on modeling the accurate temporal context of the object observations.
Experiment 3 confirms that expanding the coordinates of sparse tensors is beneficial.
Without the regression sub-head in the RoI head (Exp. 4), the performance drops validate that regressing the BBox targets in early stage is beneficial on capturing the object geometry. This can be observed in the dashed red boxes of \cref{fig:ablation} on both datasets. At last, Experiment 5 shows that the cross attention with the global RoI features can also slightly improve the performance. 

In general, the AP deviations among all ablation experiments of DairV2Xt are smaller than that of OPV2Vt, especially for the ablation with and without point-wise timestamp. We explain this from two aspects. First, the inaccurate ground-truth (see objects in \cref{fig:ablation} indicated by black arrows) of DairV2Xt weakens the general performance and lead to inaccurate evaluation. As the example shown in \cref{fig:ablation}, the full StreamLTS model has detected the unlabeled objects while the model without RoI regression has not. The wrong ground-truth will give better evaluation to the later case. Second, one sensor of DairV2xt is static and all data is captured at intersections where most vehicles are static waiting for the traffic lights. This leads to less dynamics compared to the multi CAVs scenarios of OPV2Vt which are captured under different road situations. As a result, less dynamic leads to less influence of inaccurate timestamps.

In cooperative perception, the shared data from cooperative vehicles might be outdated because of the long data processing and transmitting time. Owning to the temporal modeling with accurate point-wise timestamps, streamLTS is able to reduce the influence of this time latency. \cref{tab:ablation_latency} shows the ablation results of fusing the data with frame-wise time latency up to 200ms. Without \textit{TempFusion}, the APs drops significantly as the latency increases. However, temporal modeling with the inaccurate frame-wise timestamps (No Pt. Time) is not helpful and even misleading. Last two rows in \cref{tab:ablation_latency} show the streamLTS performance without and with latency augmentation during training. Both reduced the influence of the latency. However, the full streamLTS with train latency has the best performance, validating the effectiveness of our model on understanding the temporal context. Similar to \cref{tab:ablation}, the performance gain of point-wise temporal modeling on DairV2Xt is significantly smaller than on the OPV2Vt because of less dynamics and inaccurate ground-truth of DairV2Xt.

\begin{table}[tb]
\caption{Average Precision (AP) of TA-COOD with frame-wise time latency.}
\centering
\begin{tabular}{c|cc|cc|cc|cc|cc|cc}
\toprule
 Dataset & \multicolumn{6}{c|}{OPV2Vt}  & \multicolumn{6}{c}{DairV2Xt}\\ \cline{1-13} 
Latency & \multicolumn{2}{c}{\cellcolor[gray]{.9}0ms} & \multicolumn{2}{c}{\cellcolor[gray]{.9}100ms} & \multicolumn{2}{c|}{\cellcolor[gray]{.9}200ms}& \multicolumn{2}{c}{\cellcolor[gray]{.9}0ms} & \multicolumn{2}{c}{\cellcolor[gray]{.9}100ms} & \multicolumn{2}{c}{\cellcolor[gray]{.9}200ms}\\\cline{1-13} 

AP threshold & \multicolumn{1}{c}{0.5\quad}    &  0.7\quad   & \multicolumn{1}{c}{0.5} &  0.7 & \multicolumn{1}{c}{0.5} &  0.7& \multicolumn{1}{c}{0.5}    &  0.7   & \multicolumn{1}{c}{0.5} &  0.7 & \multicolumn{1}{c}{0.5} &  0.7\\ \hline
No TempFusion
&\multicolumn{1}{c}{79.2} & 56.1 & \multicolumn{1}{c}{74.0} & 40.9 & \multicolumn{1}{c}{60.8} & 34.0 
&\multicolumn{1}{c}{58.6} & 38.9 & 54.3 & 35.1 & 51.2 & 33.1\\ 
No Pt. Time
& \multicolumn{1}{c}{73.4} & 38.7 & \multicolumn{1}{c}{69.4} & 32.7 & 65.4 & 33.7
& \multicolumn{1}{c}{59.0} & 33.5 & \multicolumn{1}{c}{57.8} & 33.2 & 57.1 & 32.6\\ 
No Train Latency  
& \multicolumn{1}{c}{83.2} & 64.0 & \multicolumn{1}{c}{76.4} & 59.0 & 74.4 & 57.9
& \multicolumn{1}{c}{59.3} & 36.2 & \multicolumn{1}{c}{55.2} & 33.3 & 53.0 & 32.3\\  
Full steamLTS
& \multicolumn{1}{c}{\textbf{85.3}} & \textbf{72.1} & \multicolumn{1}{c}{\textbf{81.6}} & \textbf{68.0} & \textbf{78.7} & \textbf{64.7}
& \multicolumn{1}{c}{\textbf{64.2}} & \textbf{40.4} & \multicolumn{1}{c}{\textbf{61.3}} & \textbf{37.9} & \textbf{59.0} & \textbf{36.4}\\ 
\bottomrule
\end{tabular}
\label{tab:ablation_latency}
\end{table}

\section{Conclusion}
In this paper, we proposed an efficient fully sparse and query-based temporal and spatial fusion framework, StreamLTS, to fuse the asynchronized data for cooperative object detection. With experiments on the OPV2Vt and DairV2Xt datasets that are specially prepared for TA-COOD, we found that temporal modeling with fine-grained point-wise timestamps can capture the temporal context more accurately and improve the predictability of the object movement. 
Comparing to the baseline models, our fully sparse framework has shown its high efficiency during training. 
For future work, we hope our findings could be inspiring for more efficient and accurate cooperative end-to-end trajectory prediction.

\par\vfill\par

\section*{Acknowledgements}
This work is supported by the projects DFG RTC1931 SocialCars.

%
%
\bibliographystyle{splncs04}
\bibliography{main}
\section*{Appendix A: OPV2Vt Generation}\label{appendA}
The overall process of generating OPV2Vt dataset is described in \cref{alg:genOPV2Vt}. This algorithm is applied to each scene of the OPV2V dataset. Note that the coordinate-related parameters $x, y, z, \theta$ are represented in the global coordinate system, where $\theta$ represents the yaw angle. The roll and pitch angles are set to $0$. Additionally, note that $N_s < N_v$, and each sensor is installed on the roof of a vehicle in the set $\mathbf{v}$.
The functions used in \cref{alg:OPV2Vt} are defined as follows: 

\textit{Interpolation$(v1, v2, n)$}:
Given the vehicle parameters $v1$ and $v2$ from two adjacent frames and the subframe index $1\leq n \leq 10$, we interpolate the coordinates $x, y, z$ and the orientation $\theta \in [0, 2\pi)$ using  \cref{eq:interp}. For the orientation, we additionally use \cref{eq:angle_correction} to resolve the direction flipping issue after interpolation.
\begin{align}
    u &= u1 + (u2 - u1) \cdot \frac{n}{10}\label{eq:interp}\\
    \theta &= \begin{cases}
        \theta - 2\pi \quad\text{ if } \theta > \pi\\
        \theta + 2\pi \quad\text{ if } \theta < -\pi
    \end{cases}\label{eq:angle_correction}
\end{align}

\textit{CarlaPlay$(\mathbf{v}, \mathbf{s}, t)$}:
Place all vehicles $\mathbf{v}$ and LiDAR sensors $\mathbf{s}$ in the CARLA simulator according to their parameters and generate the corresponding point cloud for the time interval $\delta t$. Note that we do not change the sensor recording frequency; each full scan requires $0.1$ seconds. Therefore, each subframe only generates data for $\delta t=0.01$ seconds, which corresponds to the scan points in a $36^\circ$ sector as shown in \cref{fig:opv2vt} (each sector is illustrated with a different color).

\textit{RandomInteger$(a, b)$}: Select one integer in the range $[a, b]$.
\newpage
\begin{algorithm}[H]
\label{alg:genOPV2Vt}
\caption{OPV2Vt Generation}\label{alg:OPV2Vt}
\begin{algorithmic}[1]
\State \textbf{Input:} OPV2V meta info $\mathcal{F}$\\
$\Delta t$ \Comment{Time interval between two adjacent frames}\\
$\mathcal{F}=\{(\mathbf{v}_i, \mathbf{s}_i, t_i) | i=\{1, 2, \dots, N_f\}\}$, \Comment{Frame data $f_i$ at all timestamps $t_i$}\\
$\mathbf{v}=\{v_{i}|i=\{1, 2, \dots, N_v\}\}$ , \Comment{Vehicle info in the i-th frame}\\
$\mathbf{s}=\{s_{i}|i=\{1, 2, \dots, N_s\}\}$ , \Comment{Sensor info in the i-th frame}\\
$v = (id, x, y, z, h, w, l, \theta)$ \Comment{BBox parameters of the j-th vehicle}\\
$s = (id, x, y, z, \theta)$ \Comment{Sensor parameters of the j-th vehicle}
\State \textbf{Output:} Generated dataset $\mathcal{D}$ 
\State \textbf{Initialize:} $\mathcal{G}=\emptyset$, $\mathcal{D}=\emptyset$, $\delta t = \Delta t / 10$ \\
\For{$i \gets 2$ to $N_f$} \Comment{Loop over all frames}
    \For{$k \gets 1$ to $10$} \Comment{Interpolate adjacent two frames into 10 subframes}
        \State $\mathbf{v}=\mathbf{s}=\emptyset$ 
        \For{$j \gets 1$ to $N_v$} \Comment{Interpolate over the parameters of each vehicle}
            \If{$v_{i-1,j}$ exists}
                    \State $\mathbf{v}\gets \mathbf{v} + \text{Interpolate}(v_{i-1,j}, v_{ij}, n$)
            \EndIf
        \EndFor
        \For{$j \gets 1$ to $N_s$} \Comment{Interpolate over the parameters of each vehicle}
            \If{$s_{i-1,j}$ exists}
                \State $\mathbf{s} \gets \mathbf{s} + \text{Interpolate}(s_{i-1,j}, s_{ij}, n$)
            \EndIf
        \EndFor
        \State $t = t_i + k\cdot\delta t$ \Comment{New timestamp for current subframe}
        \State $\mathcal{G} \gets \mathcal{G} + (\mathbf{v}, \mathbf{s}, t)$ \Comment{Save data for current subframe}
    \EndFor
\EndFor
\State $\mathcal{P}=\emptyset$
\For{$(\mathbf{v}, \mathbf{s}, t)$ in $\mathcal{G}$} \Comment{Simulation replay over all new subframes}
    \State $\mathbf{p} \gets$ CarlaPlay$(\mathbf{v}, \mathbf{s}, t)$ \Comment{Generate $1/10$ lidar scans for each vehicle in $g$}
    \State $\mathcal{P} \gets \mathcal{P}+(\mathbf{p}, t)$ \Comment{Save scanned point clouds}
\EndFor
\State $\mathcal{P} \gets \text{Sort}_{t}(\mathcal{P})$ \Comment{Sort data in ascending order of $t$}
\State $t_\text{max} = \max ({t|(\mathbf{p}, t)\in \mathcal{P}})$
\For{$j \gets 1$ to $N_s$} \Comment{Compose full scans with random sensor ticking times}
    \State $\tau = \text{RandomInteger}(1, 5) \cdot \delta t$ \Comment{Generate random sensor ticking time}
    \While{$\tau < t_\text{max}$}
        \State $P \gets \{p_j|(p_j, s_j)\in \mathbf{p}, (\mathbf{p}, t) \in \mathcal{P}, \tau \leq t <\tau + \Delta t\}$ \Comment{Full scan}
        \State $s = s_j$ where $(p_j, s_j)\in \mathbf{p}, (\mathbf{p}, t) \in \mathcal{P}, t=\tau$ \Comment{Sensor pose}
        \State $G \gets \mathbf{v} \text{ where } (\mathbf{v}, \mathbf{s}, \tau + \Delta t) \in \mathcal{G}$ \Comment{Ground-truth BBoxes}
        \State $\mathcal{D} \gets \mathcal{D} + (P, G, s, \tau)$
        \State $\tau \gets \tau + \Delta t$
    \EndWhile
\EndFor
\end{algorithmic}
\end{algorithm}

\section*{Appendix B: DairV2Xt Generation}\label{appendB}

The overall process of generating DairV2Xt dataset is described in \cref{alg:genDairV2Xt}. Note that this algorithm is applied to each sequence containing $N_f$ frames. The functions used in \cref{alg:genDairV2Xt} are defined as follows: 

\textit{Register$(O, R)$}: Given the reference point cloud $O$ and the source point cloud $R$, this function registers $R$ to $O$. The Iterative Closest Point (ICP) algorithm is used for the registration with a threshold distance of $0.2$m. Note that during registration, dynamic points are moved via the BBox annotations. The registration result is a transformation matrix $M$, which is then used to align the point cloud $R$ to $O$. The transformed $R$ is merged into $O$, and the merged point cloud is downsampled with a voxel size of $0.1$m to speed up the registration process. The \textit{Register} function returns the updated reference point cloud $O$ and the sensor parameters $s$ obtained from $M$.

\textit{Transform$(\mathbf{v}, s)$}: This function constructs a transformation matrix from $s$ and then transforms the parameters of each $v \in \mathbf{v}$ using the matrix. It returns the transformed bounding box (BBox) parameters.

\textit{RetriveTimestamp(v, R)}: Given the BBox parameters $v$ and the point cloud $R$, this function finds the points $P = \{p \mid p \in R\}$ that lie within the BBox $v$ and returns the mean timestamp of the points in $P$.

\textit{Track$(\mathcal{T}, \mathbf{u})$}: The input $\mathcal{T}$ contains $N$ trajectories, where each trajectory $T_i$ is a list of BBoxes $[v_1, v_2, \dots, v_j, \dots, v_n]$ with $v_j = ((x, y, z, h, w, l, \theta), t)$. Each input BBox $u_k \in \mathbf{u}$ has the same parameterization as $v_j$. For each $u_k$, the Euclidean distance $d_{i}$ is calculated between $u_k$ and each $v_n \in \mathcal{T}$. If $d_i < 3$m, $u_k$ is appended to $T_i$; otherwise, a new trajectory $T_{N+1} = [u_k]$ is initialized and added to $\mathcal{T}$. The updated tracklets $\mathcal{T}$ are then returned.

\textit{Interpolate$(\mathcal{T}, t)$}: For each trajectory $T_i$ in $\mathcal{T}$, the function finds the two BBoxes $b1 = (v1, t1)$ and $b2 = (v2, t2)$ that are closest in time and fulfill the condition $t1 < t \leq t2$. If such BBoxes exist, interpolation is performed using
\begin{equation}
    u = u1 + (u2 - u1) \cdot \frac{t - t1}{t2 - t1}
\end{equation}
for the location $x, y, z$ and the orientation $\theta$ of the BBox. The mean dimensions $h, w, l$ of the two BBoxes $v1$ and $v2$ are used for the new interpolated BBox.

\newpage
\begin{algorithm}[H]
\caption{DairV2Xt  Generation}\label{alg:genDairV2Xt}
\begin{algorithmic}[2]
\State \textbf{Input:} DairV2X dataset $\mathcal{F}$\\
$\Delta t$ \Comment{Time interval between two adjacent frames}\\
$\mathcal{F}=\{(P_i, Q_i, \mathbf{v}^P_i, \mathbf{v}^Q_i, s^P_i, s^Q_i, t_i) | i\in\{1, 2, \dots, N_f\}\}$, \Comment{Frame data at all times $t_i$}\\
$P = \{x_i, y_i, z_i, t_i|i\in\{1, 2, \dots, N_P\}\}$, \Comment{CAV Point cloud in each frame}\\
$Q = \{x_i, y_i, z_i, t_i|i\in\{1, 2, \dots, N_Q\}\}$, \Comment{Infrastructure Point cloud in each frame}\\
$\mathbf{v}^*=\{v_{i}|i=\{1, 2, \dots, N_v\}\}$ , \Comment{BBoxes in point cloud $*$}\\
$s^* = (x, y, z, \theta)$ \Comment{Sensor parameters for point cloud $*$}\\
$v = (x, y, z, h, w, l, \theta)$ \Comment{BBox parameters}
\State \textbf{Output:} Generated dataset $\mathcal{D}$ 
\State \textbf{Initialize:} $\mathcal{D}=\emptyset$ \\
\State $idx \gets \argmin_i dist(s^P_{i}, s^Q_{i})$ \Comment{Index of min. norm-2 distance of sensor positions}
\State $O_\text{ref} \gets P_{idx}$ \Comment{Global reference frame}
\For{$i \gets idx$ to $N_f$} \Comment{Register $P$ for $i>idx$}
    \State $(O_\text{ref}, s^P_i) \gets \text{Register}(O_\text{ref}, P_i)$
\EndFor
\For{$i \gets idx$ to $1$} \Comment{Register $P$ for $i<idx$}
    \State $(O_\text{ref}, s^P_i) \gets \text{Register}(O_\text{ref}, P_i)$
\EndFor

\For{$i \gets 1$ to $N_f$}
    \State $(\cdot, s^Q_i) \gets \text{Register}(O_\text{ref}, Q_i)$ \Comment{Register $Q$ }
    \State $\mathbf{v}^P_i \gets \text{Transform}(\mathbf{v}^P_i, s^P_i)$ \Comment{Transform BBoxes of $P$ to aligned frame $O_\text{ref}$}
    \State $\mathbf{v}^Q_i \gets \text{Transform}(\mathbf{v}^Q_i, s^Q_i)$ \Comment{Transform BBoxes of $Q$ to aligned frame $O_\text{ref}$}
    \For{$(\mathbf{v}, R_i) \text{ in } \{(\mathbf{v}^P_i, P_i), (\mathbf{v}^Q_i, Q_i)\}$ }
        \State $\mathbf{u} \gets \emptyset$
        \For{$v$ in $\mathbf{v} $} \Comment{Timestamping each BBox}
            \State $t \gets \text{RetriveTimestamp}(v, R_i)$
            \State $\mathbf{u} \gets \mathbf{u} + (t, v)$       
        \EndFor
        \State $\mathcal{T} \gets \text{Track}(\mathcal{T}, \mathbf{u})$ \Comment{Track BBoxes}
    \EndFor
\EndFor
\For{$i \gets 1$ to $N_f$}
    \State $t \gets \max_t (\{t|(x, y, z, t) \in P_i\})$ \Comment{Align global timestamp to maximum $t$ in $P$}
    \State $\mathbf{v} \gets \text{Interpolate}(\mathcal{T}, t)$ \Comment{Get BBoxes at time $t$}
    \State $\mathcal{D} \gets \mathcal{D} + (P_i, Q_i, s^P_i, s^Q_i, \mathbf{v})$ \Comment{Save data for frame $i$}
\EndFor
\end{algorithmic}
\end{algorithm}

\end{document}